\documentclass[11pt]{article}

\newif\ifarxiv
\arxivtrue

\newif\ifreviewmode
\reviewmodefalse %

\renewcommand{\baselinestretch}{0.99}

\ifreviewmode
\usepackage[review]{acl}
\else
\usepackage{acl}
\fi

\usepackage{times}
\usepackage{latexsym}

\usepackage[T1]{fontenc}

\usepackage[utf8]{inputenc}

\usepackage{microtype}
\usepackage{inconsolata}

\usepackage{amsmath,amsfonts,bm}

\def\eqref#1{equation~\ref{#1}}

\def\1{\bm{1}}

\DeclareMathAlphabet{\mathsfit}{\encodingdefault}{\sfdefault}{m}{sl}
\SetMathAlphabet{\mathsfit}{bold}{\encodingdefault}{\sfdefault}{bx}{n}

\newcommand{\R}{\mathbb{R}}

\usepackage{mathtools}
\usepackage{booktabs}
\usepackage{xspace}
\usepackage{tipa}

\usepackage{xcolor}

\usepackage{wrapfig}

\usepackage{caption}
\usepackage{subcaption}

\usepackage{algpseudocode}
\usepackage{algorithm}

\definecolor{caribbeangreen}{rgb}{0.0, 0.8, 0.6}
\definecolor{carrotorange}{rgb}{1.0, 0.615, 0.0}

\newcommand{\trans}[2]{\ensuremath{\xrightarrow{#1 \, : \, #2}}}
\newcommand{\transb}[2]{\ensuremath{\xrightarrow{\textbf{\texttt{#1}} \, : \, \textbf{\texttt{#2}}}}}

\newcommand{\method}[0]{SIP\xspace}
\newcommand{\methodexplain}[0]{\textbf{S}imulation-\textbf{I}nduced \textbf{P}rior\xspace}

\newcommand{\grayout}[1]{\textcolor{gray}{#1}}

\newcommand{\eg}{e.g.~}
\newcommand{\ie}{i.e.~}

\usepackage{multicol}
\usepackage{multirow}

\newcommand{\dfourmethod}{\method-d4\xspace}
\renewcommand{\paragraph}[1]{\textbf{#1.}\xspace}

\newcommand{\added}[1]{#1}

\usepackage{enumitem}
\setlist[itemize]{parsep=0pt,topsep=2pt,itemsep=0ex,partopsep=1ex,parsep=1ex, leftmargin=2ex}

\newcommand{\tightmath}{\abovedisplayskip=5pt
\belowdisplayskip=5pt}

\usepackage{hyperref}

\usepackage{url}
\usepackage[capitalise]{cleveref}

\usepackage{graphicx}

\title{SIP: Injecting a Structural Inductive Bias \\ into a Seq2Seq Model by Simulation}

\author{Matthias Lindemann$^{\includegraphics[scale=0.4]{images/teapot\_color.pdf}}$ \and Alexander Koller$^{\includegraphics[scale=0.43]{images/beverage\_box\_color.pdf}}$ \and Ivan Titov$^{\includegraphics[scale=0.42]{images/teapot\_color.pdf}, \includegraphics[scale=0.38]{images/glass\_of\_milk\_color.pdf}}$ \\
	$^{\includegraphics[scale=0.4]{images/teapot\_color.pdf}}$ILCC, University of Edinburgh, \hfill
	$^{\includegraphics[scale=0.43]{images/beverage\_box\_color.pdf}}$LST, Saarland University, \hfill
	$^{\includegraphics[scale=0.38]{images/glass\_of\_milk\_color.pdf}}$ILLC, University of Amsterdam \\
	{\small \texttt{m.m.lindemann@sms.ed.ac.uk}, \texttt{koller@coli.uni-saarland.de}, \texttt{ititov@inf.ed.ac.uk} }
}

\begin{document}

\maketitle

\begin{abstract}
Strong inductive biases enable learning from little data and help generalization outside of the training distribution.
Popular neural architectures such as Transformers lack strong structural inductive biases for seq2seq NLP tasks on their own. Consequently, they struggle with systematic generalization beyond the training distribution, \eg with extrapolating to longer inputs, even when pre-trained on large amounts of text.
We show how a structural inductive bias can be efficiently injected into a seq2seq model by pre-training it to simulate structural transformations on synthetic data.
Specifically, we inject an inductive bias towards Finite State Transducers (FSTs) into a Transformer by pre-training it to simulate FSTs given their descriptions.
Our experiments show that our method imparts the desired inductive bias, resulting in improved systematic generalization and better few-shot learning for FST-like tasks. 
Our analysis shows that fine-tuned models accurately capture the state dynamics of the unseen underlying FSTs, suggesting that the simulation process is internalized by the fine-tuned model.%
\ifarxiv %
\footnote{
	We release our code at \href{https://github.com/namednil/sip}{https://github.com/namednil/sip}}
\else
\fi
\end{abstract}

\vspace{-6pt}
\section{Introduction}

Inductive biases, \ie the preferences and the abstract knowledge a model brings to the task, enable a model to learn from small amounts of data and generalize systematically outside of the training distribution. While seq2seq models perform very well on in-distribution data, they usually lack structural inductive biases and consequently struggle with systematic generalization. Previous work has shown that this includes generalization to unseen combinations of known sub-strings \citep{lake2018generalization, keysers2020measuring}, extrapolation to longer inputs \citep{hupkes2020compositionality} and deeper recursion \citep{kim-linzen-2020-cogs}.

Integrating structural inductive biases into seq2seq models is challenging. One popular approach is to develop specialized architectures \citep{wu-cotterell-2019-exact, zheng-lapata-2021-compositional-generalization, kim-2021-nqscfg, lindemann-etal-2023-compositional-generalization}, which makes it difficult to precisely control and adjust the nature of the inductive bias to changing demands as the architecture would need to be modified and models re-trained. Recently, some works instead have tried to inject inductive biases into seq2seq models by pre-training on a well-chosen synthetic task \citep{krishna-etal-2021-pretraining-summarization, wu2021lime, wu2022insights} or meta-learning on a distribution of synthetic tasks \citep{mccoy2020universal, mccoy2023modeling} using MAML \citep{finn2017model}. Here, the inductive bias can be controlled by the choice of the synthetic task. However, meta-learning with MAML scales poorly because it requires expensive second-order derivatives and standard pre-training can be less effective \citep{mccoy2023modeling}.

\begin{figure}[t]
    \centering
    \includegraphics[width=\linewidth]{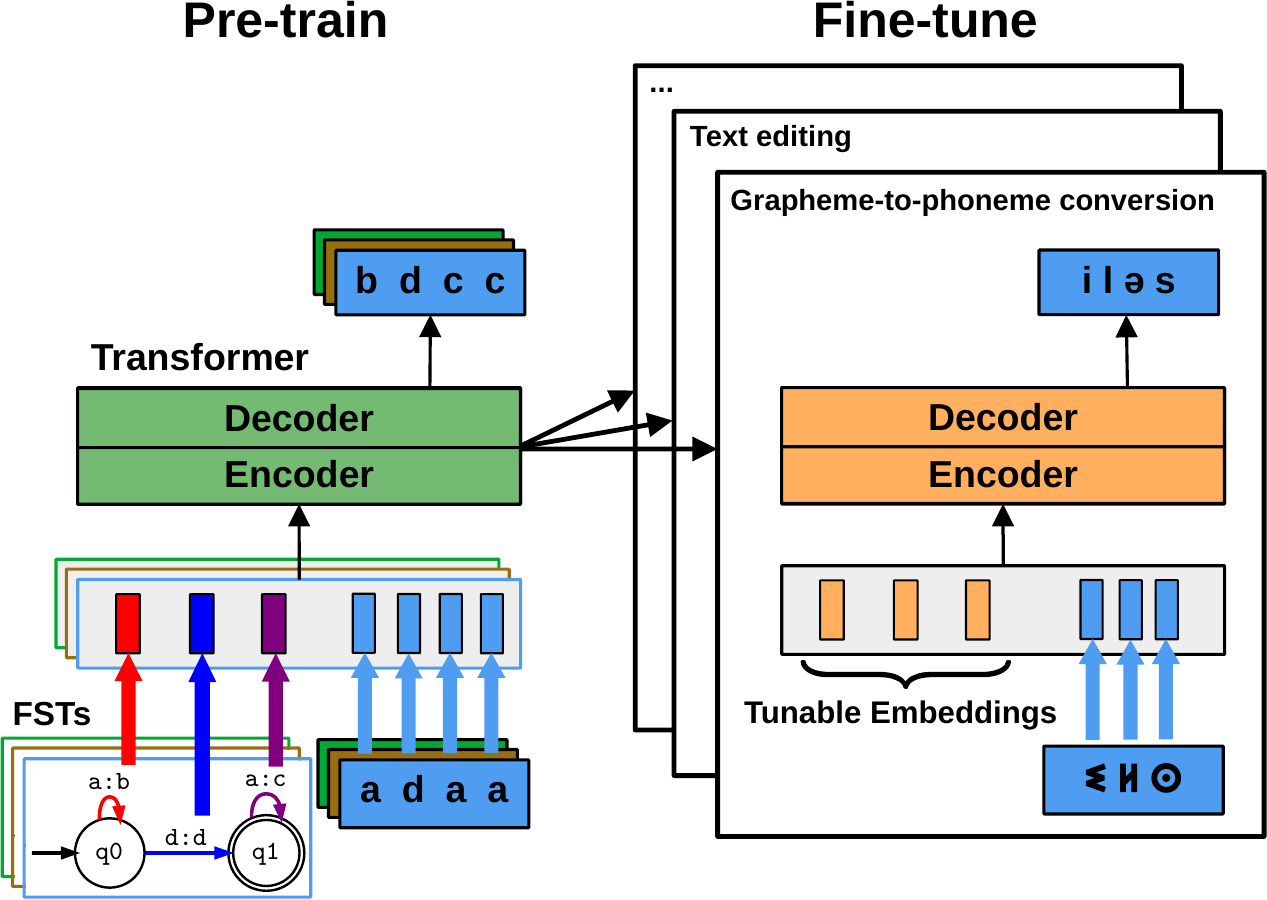}
    \caption{Left: Pre-training a Transformer to simulate automatically generated FSTs. 
    Right: fine-tuning the Transformer and the prefix where the FST used to be on a downstream task by using only input/output pairs. Tunable parameters are represented in orange.}
    \label{fig:overview}
    \vspace{-12pt}
\end{figure}

In this work, we present a computationally inexpensive way of injecting a structural inductive bias into a Transformer. We focus specifically on introducing an inductive bias that is helpful for tasks that traditionally have been approached with Finite State Transducers (FSTs). We choose FSTs because they are formally well understood, are easy to generate automatically, and are one of the simplest computational devices that are useful in NLP applications. While we focus on FSTs, the methodology is fairly general and our approach also provides a starting point for incorporating more general structural biases, provided by more expressive formalisms such as Pushdown Transducers.

Our approach (\method, for \methodexplain) is simple (see \cref{fig:overview}): given a representation of an FST and an input string, a Transformer is pre-trained to predict what the output of the FST is on the given input. 
We assume that FSTs are not specified for fine-tuning on downstream tasks, so we replace the FST with tunable embeddings and fine-tune the model solely on input/output pairs. Since we fine-tune all parameters, the model can deviate from FST-like behavior if needed.

\paragraph{Contributions} 
We show that a model pre-trained with \method has an inductive bias that improves systematic generalization and few-shot learning for `FST-like' downstream tasks. \method not only improves systematic generalization on FST tasks similar to those seen during pre-training but also on ones that are structurally more complex. The same pre-trained model also transfers well to natural data and achieves strong results on few-shot learning of text editing (\eg Jane Doe $\rightarrow$ J. Doe) and grapheme-to-phoneme conversion.

Our probing experiments give insights into how the inductive bias is injected: \method not only leads to the imitation of the input/output behaviour of FSTs, but encourages dynamics to emerge that \textit{simulate} crucial aspects of FSTs in the hidden representations. Fine-tuning can leverage these dynamics, providing the inductive bias, and learn representations that resemble those of ground truth FSTs.

\section{Related Work}

\paragraph{Systematic generalization} Systematic generalization refers to the ability of a model to generalize (or extrapolate) beyond its training distribution in a systematic way that aligns with how humans generalize. Systematic generalization is difficult for standard seq2seq models in contexts such as semantic parsing \citep{finegan-dollak-etal-2018-improving} and machine translation \citep{li-etal-2021-compositional}, in particular to unseen combinations of phrases, longer inputs as well as deeper recursion \citep{keysers2020measuring, kim-linzen-2020-cogs}.

A range of approaches have been developed to tackle this, with many works focusing on specialized architectures \citep{guo2020hierarchical, kim-2021-nqscfg, lindemann-etal-2023-compositional-generalization}. \citet{furrer2020compositional} find that the specialized architectures they consider do not transfer well to tasks beyond the context in which they were designed. This highlights the importance of being able to adjust inductive biases more easily than re-designing the architecture of a model.
Large-scale pre-training has also been shown to help with systematic generalization \citep{furrer2020compositional}. However, challenges remain even for LLMs such as GPT-3 and PALM \citep{qiu-etal-2022-evaluating, dziri2023faith}. 
The methodology we present in this work can be used to create additional material for LLM pre-training. Here we focus on smaller models and leave this to future work.

\paragraph{Pre-training with synthetic tasks} Pre-training a model on a synthetic task to introduce specific inductive biases has been explored by several recent works. \citet{krishna-etal-2021-pretraining-summarization} identify useful `skills' for news summarization and develop a pre-training task accordingly. LIME \citep{wu2021lime} targets mathematical reasoning and is pre-trained on string manipulation that resembles formal reasoning. %
\citet{papadimitriou2023pretrain} consider pre-training with several synthetic languages to investigate which helps most for language modelling. In contrast to these works, our approach targets simulating a computational device and maintains a closer relation to the pre-training setting because of the tunable prefix.

A challenge for using individually hand-crafted tasks is to cover a sufficient space of phenomena that are relevant to downstream tasks. Instead of training on a single task only, \citet{mccoy2020universal, mccoy2023modeling} meta-learn on a distribution of tasks using MAML \citep{finn2017model}. 
Our approach also uses a distribution of tasks but it scales better than MAML-based methods because MAML requires computing and storing second-order derivatives. For example, the Transformer we train has a magnitude more parameters than the LSTM of \citet{mccoy2023modeling} and is pre-trained on a smaller GPU (A100 vs RTX 2080 TI).
In addition, as the complexity of each individual task grows, MAML requires more examples per task. We circumvent this by using a compact and unambiguous description of each task instead.

\paragraph{Simulating execution} The idea of using a neural network to predict the outcome of the execution of a computational device or code has come up in several contexts over the last few years. Early work by \citet{zaremba2014learning} investigates it as a challenging benchmark for LSTM-based seq2seq models. Recent works have explored simulating (aspects of) code execution for various down-stream applications, such as program synthesis \citep{austin2021program}, or debugging and code analysis \citep{bieber2022static} as well as reverse engineering \citep{10.1145/3468264.3468607}. 
Finally, \citet{finlayson-etal-2022-makes} train a Transformer to interpret regular expressions: given a regular expression and a string, the task is to decide if the string is in the regular language. There are three crucial differences between their work and ours: (i) they investigate the empirical capabilities of Transformers while we introduce structural inductive biases for downstream tasks, (ii) they consider binary outputs whereas we consider sequential outputs, and (iii) we perform probing experiments showing strong evidence for FST simulation in the hidden representations.

\paragraph{Emergent World Representations} Our analysis provides evidence that our model trained with \method internally simulates transitions between FST states even though it was not explicitly supervised to do so. Similar observations have been made for Language Models trained to play Othello \citep{li2023emergent} and chess \citep{karvonen2024emergent}, where the model was found to acquire a representation of the board state simply from being trained to predict the next move.

\section{Finite State Transducers}
\label{sec:background}

We briefly review Finite State Transducers (FSTs) which we use in our experiments.
FSTs are closely related to Finite State Automata (FSAs). While an FSA describes a set of strings, an FST describes a \textit{relation} between strings, \ie a set of pairs $(x,y)$, where $x$ is an input $y$ is an output.

\begin{figure}[t]
\centering
     \begin{subfigure}[t]{0.39\linewidth}
         \centering
         \includegraphics[width=\linewidth]{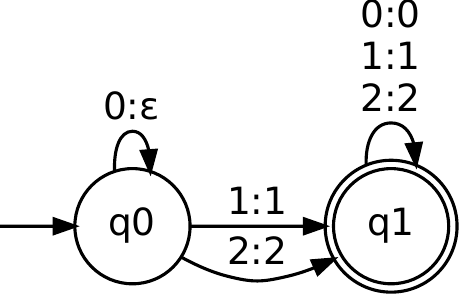}
         \caption{A deterministic FST.}
         \label{fig:deterministic-fst}
     \end{subfigure}
     \hfill
     \begin{subfigure}[t]{0.565\linewidth}
         \centering
         \includegraphics[width=\linewidth]{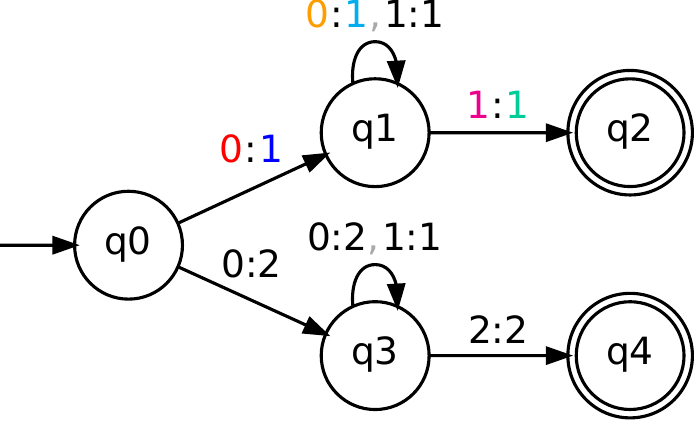}
         \caption{A non-deterministic but functional FST.}
         \label{fig:functional-fst}
     \end{subfigure}
    \caption{Examples of \textit{functional} FSTs. The FST in (a) deletes leading zeros. The FST in (b) replaces any \texttt{0} by a \texttt{1} if the last input symbol is a \texttt{1}. Conversely, if the last symbol is a \texttt{2}, any \texttt{0} is replaced by a \texttt{2}. The output can only be determined after the last input symbol.}
    \label{fig:examples-fst}
    \vspace{-12pt}
\end{figure}

FSTs can be visualized as labelled directed graphs (see \cref{fig:examples-fst}), where the nodes are called \textit{states} and the edges are called \textit{transitions}. 
Consider the path $\texttt{q0} \transb{\textcolor{red}{0}}{\textcolor{blue}{1}} \texttt{q1} \transb{\textcolor{carrotorange}{0}}{\textcolor{cyan}{1}} \texttt{q1} \transb{\textcolor{magenta}{1}}{\textcolor{caribbeangreen}{1}} \texttt{q2}$ in \cref{fig:functional-fst}.
This path is called an \textit{accepting path} because it starts in an \textit{initial} state (indicated by an arrow `from nowhere' pointing to the state), and it ends in a \textit{final} state (indicated by double circles). An accepting path shows what an input can be mapped to. In this case, the path shows that the FST transduces the input \texttt{\textcolor{red}{0}\textcolor{carrotorange}{0}\textcolor{magenta}{1}} into the output \texttt{\textcolor{blue}{1}\textcolor{cyan}{1}\textcolor{caribbeangreen}{1}}. We can read off which input an accepting path associates an output to by concatenating all the strings along the path occurring before `:'. The output can be determined by concatenating the strings after `:'. Hence, each transition \trans{\sigma}{\rho} can be thought of as `replacing' $\sigma$ by $\rho$. Inserting and deleting can be achieved by means of the empty string, written as $\epsilon$. For example, \cref{fig:deterministic-fst} `replaces' leading zeros by an empty string, effectively deleting them. 

In general, an input can be paired with arbitrarily many different outputs. We call an FST $f$ \textbf{functional} if every input $x$ is paired with at most one output $y$, and use the notation $f(x)$ to refer to $y$. All FSTs we consider here are functional.

In this work, we investigate generalization across different sub-classes of FSTs, namely from the less expressive deterministic FSTs to non-deterministic FSTs.
An FST is called \textbf{deterministic} if (i) it has a unique initial state, (ii) for all states $q$ and input symbols $\sigma$ there is at most one transition $q \trans{\sigma}{\rho} q'$ and (iii) $\sigma \neq \epsilon$. 
Intuitively, this means that in any state, for an input symbol $\sigma$ there is at most one possible next state and one possible output, and hence for any input string there is at most one path that is compatible with it. Because of this, we can always infer a prefix of the output by looking only at a \textit{prefix} of the input string and ignoring the rest. 
For example, consider the input prefix \texttt{001}. In the deterministic FST in \cref{fig:deterministic-fst}, we know that the output has to start with \texttt{1} because there is only one path that is compatible with \texttt{001}. In contrast, in the non-deterministic FST in \cref{fig:functional-fst}, three paths are compatible with \texttt{001} that have different outputs. In that case, we can only determine the output once we look at the last symbol of the input. In short, non-deterministic FSTs can take context to the right into account but deterministic FSTs cannot.

\section{Simulation-Induced Prior}
\label{sec:prior-dist}

Our approach largely follows the pre-training and fine-tuning paradigm. We first pre-train on synthetic FST tasks by giving the model a representation of an FST as a prefix and an input string (see \cref{fig:overview}). The training objective is to predict the output of the FST on the input string.
Our research hypothesis is that training a model to predict the behaviour of an FST incentivizes the model to acquire reusable dynamics that internally simulate FSTs.
When fine-tuning the model using a tunable prefix instead of an encoding of an FST, these dynamics should be easy to leverage and provide a structural inductive bias for FST-like tasks.

\subsection{Pre-training}
\label{sec:pre-train}
During pre-training, the model is given a representation of an FST and a string in its domain and has to predict the output of that FST on the given input string. The input to the Transformer is a sequence of vectors from $\R^d$, which consist of a prefix that represents the FST $f$ and a suffix comprised of the embeddings of the input string (see \cref{fig:overview}):
\tightmath
\begin{align*}
\underbrace{\mathbf{h}_1, \mathbf{h}_2,  \ldots, \mathbf{h}_k}_{\text{FST encoding}}, \underbrace{\mathbf{x}_1, \mathbf{x}_2  \ldots, \mathbf{x}_n}_{\text{Input to FST}}
\end{align*}
Each $\mathbf{h}_i$ encodes a transition $p \trans{\sigma}{\rho} q$ as a vector:
\begin{align*}
    \mathbf{h}_i = W [& \textsc{emb}_{\text{State}}(p) ; \textsc{emb}_{\text{State}}(q); \\
    & \textsc{emb}_{\text{Symbol}}(\sigma); \textsc{emb}_{\text{Symbol}}(\rho) ; \textsc{emb}_{\text{Final}}(e) ]
\end{align*}
where $[;]$ represents vector concatenation, $e$ indicates if $q$ is a final state, and $W$ is linear layer that ensures that $\mathbf{h} \in \R^d$. All embeddings are simple look-up tables based on the id of the state or symbol.
The initial state of the FST is always assigned the id 0, and positional embeddings are used as usual.
The model is trained to maximize the log probability of the output $y=f(x)$ of the FST $f$.

\subsection{Fine-tuning}
\label{sec:fine-tune}
After pre-training, we can apply our model to a downstream task and fine-tune it. We assume we do not have access to an FST for the downstream task, and therefore we replace the FST encoding with a sequence of tunable embeddings. That is, the input to the model is a sequence of vectors:
\begin{align*}
\underbrace{\mathbf{h'}_1, \mathbf{h'}_2,  \ldots, \mathbf{h'}_k}_{\text{Tunable embeddings}}, \underbrace{\mathbf{x}_1, \mathbf{x}_2  \ldots, \mathbf{x}_n}_{\text{Input}}
\end{align*}
where $\mathbf{x}_1, \mathbf{x}_2  \ldots, \mathbf{x}_n$ are the embeddings of the input tokens, $\mathbf{h'}_i \in \mathbb{R}^d$ are the tunable embeddings and $k$ is a hyperparameter.
The embeddings $\mathbf{h'}_i$ are initialized to the average of the encoding of multiple FSTs from the pre-training phase. 
The most straightforward way to fine-tune is to only modify $\mathbf{h'}$ because we are looking for an FST-like task representation. This is similar to prompt tuning \citep{lester-etal-2021-power}. However, this does not work well on tasks outside the pre-training distribution. Hence, we fine-tune the entire model%
, including the prefix, and use a higher learning rate for the prefix than for the rest of the model (see \cref{appendix:additional-model-details}).

\subsection{Constructing Pre-Training Data}
\label{sec:construct-pretrain}
To create our pre-training data, we sample 40,000 deterministic FSTs. For every FST, we sample 5 input/output pairs with input lengths up to 35. In total, this leads to 200,000 pairs for training along with their FSTs.
To describe the sampling procedure in more detail, we use an overall vocabulary $V$ consisting of the printable ASCII tokens and the Unicode block for IPA symbols (used for transcribing speech). Seq2seq tasks in the wild usually do not use the whole space of this vocabulary, so for each task $T$ we first uniformly sample the vocabulary size $|V_T|$ between 5 and 25 and then uniformly select a subset $V_T \subseteq V$. Then, we uniformly sample the number of states $|Q_T|$ between 2 and 4, and the number of final states between 1 and $|Q_T|$. 
For every state $q$ and every symbol $\sigma \in V_T$ we introduce at most one outgoing transition to a state $q'$, chosen uniformly at random. This ensures that the FST is deterministic. We then sample the output for the transition: either a symbol $\rho \in V_T$ or $\epsilon$. Finally, we minimize the number of states of the FST using OpenFST \citep{allauzen2007openfst}, and exclude those without cycles, as they express finite relations. See \cref{appendix:deterministic-fsts} for details.

In practical applications of FSTs, in particular for text editing, one often wants to keep certain parts of the input unchanged. This can be achieved with a set of transitions of the form $q \trans{\sigma}{\sigma} q'$ for all $\sigma \in V_T$. Since it is very unlikely to sample such a set of transitions, we use a special symbol that acts as a shorthand for this, which we also use when encoding the FST for pre-training.

\section{Evaluating \method's Inductive Bias}
\label{sec:evaluating-inductive-bias}

To understand the effects of our pre-training procedure on the inductive bias of the model and on the downstream performance, we first explore systematic generalization on synthetic FST tasks. This allows us to precisely control the similarity between the pre-training and the downstream task.

\subsection{Evaluation Methodology}

To evaluate the degree to which a model has an inductive bias towards FSTs, we now describe two methods for generating training and test data that reward a model for showing important aspects of FST-like systematic generalization.

\paragraph{Iteration generalization} Cycles are a characteristic feature of FSTs, and iteration generalization tests if a model learns that cycles can be traversed more often than seen during training. More specifically, given an FST, we generate training data which requires visiting any state only a few times (\textit{iteration count} up to 3). In the test data, the model has to generalize to visiting states more often (iteration count at least 4). This notion is related to length generalization \citep{lake2018generalization} but tailored specifically to FSTs.

\paragraph{Unseen combinations of transitions (UC)}
\label{sec:UC}
When an FST processes a string, the set of possible next transitions only depends on the current FST state; it does not matter how the current state was reached. 
Hence, a model with an inductive bias towards FSTs should also not be overly sensitive to how a state is reached, and correctly handle situations where a specific \textit{combination} of transitions was unobserved during training. For example, consider the FST in \cref{fig:deterministic-fst}, which deletes leading zeros from a number.
Suppose that a model is trained on examples such as \texttt{0012}, \texttt{2201}, \texttt{1012} but no training example contains the combination of leading zeros followed by a \texttt{2}, which corresponds to using the combination of the transitions $\texttt{q0}\trans{0}{\epsilon}\texttt{q0}$ and $\texttt{q0}\trans{2}{2}\texttt{q1}$. If the model has an inductive bias towards FSTs, it should generalize to this unseen combination and correctly handle inputs such as \texttt{0021}.
To generate appropriate training and test data for this, we sample a pair of adjacent transitions (such as $\texttt{q0}\trans{0}{\epsilon}\texttt{q0}$ and $\texttt{q0}\trans{2}{2}\texttt{q1}$ in \cref{fig:deterministic-fst}) and ensure that no training example uses both transitions within the \textit{same} string. In contrast, in the test data, all examples require using the \textit{combination} of the transitions. To make the generalization setup more challenging, we ensure this for multiple pairs of transitions at the same time. We refer to \cref{appendix:uc} for details on the construction.

UC is related to the method of \citet{keysers2020measuring} who also withhold combinations of seen elements to assess systematic generalization.

\subsection{Setup and Baselines}

To make a fair comparison, all models we experiment with in the main paper share the same architecture and are initialized from the same checkpoint before any additional pre-training, namely ByT5-small \citep{xue2022byt5}. ByT5 has 300 million parameters and was pre-trained on the multilingual C4 corpus. 
It uses raw bytes as tokens, which enables full Unicode support and is a natural unit to consider for FST-like tasks such as text editing and grapheme-to-phoneme conversion.
We report additional results with a T5-Base model in \cref{appendix:t5-base}, where we observe similar trends.

\paragraph{\dfourmethod} This is a model using the method we propose in this work. We pre-train on the data generated in \cref{sec:construct-pretrain} (\textbf{d}eterministic FSTs, with up to \textbf{4} states) for 20 epochs. 
This model achieves an average sequence-level accuracy of 98\% on predicting the output of an unseen FST from the training distribution. 
For fine-tuning, we use a prefix of length 50 for all experiments in this paper. 
\added{As an ablation, we also fine-tune the model without the prefix of tunable embeddings (-prefix).}

\paragraph{Naive pre-training} We use the same pre-training data as for \dfourmethod but omit the description of the FST and only train on input/output pairs. 

\paragraph{Task embeddings (TE)} TE is a simplified version of \method. Instead of using an encoding of an FST in the prefix, this baseline uses 50 randomly initialized embeddings specific to each FST. The embeddings are learned from examples jointly with the rest of the model. Several works have used a single embedding to encode a domain/task in multi-task learning \citep{tsvetkov-etal-2016-polyglot, stymne-etal-2018-parser, zhang-etal-2022-task}. Using a shorter tunable prefix resulted in considerably worse performance in our setup.
TE is fine-tuned analogously to \method, \ie with a prefix of tunable embeddings.

\paragraph{Set} \citet{wu2022insights} investigate the effectiveness of 18 simple synthetic pre-training tasks and found Set to perform best on average. 
The task is to deduplicate characters such that every type occurs only once, e.g. the input \texttt{dabacd} becomes \texttt{dabc}. This task can be represented by a deterministic FST, albeit a very large one with $2^n$ states for a vocabulary of size $n$.

\subsection{Systematic Generalization within the Pre-training Distribution}
\label{sec:within-pretrain}

First, we want to establish to what degree the pre-training has conferred any inductive bias on the distribution it was pre-trained on.

\paragraph{Setup} For each generalization setup, we generate 5 unseen FSTs with 4 states each using the same procedure as for the pre-training.
We fix the vocabulary size to its maximum value (25) in the pre-training data and only use printable ASCII characters in order to reduce variance across tasks.
To evaluate UC, we withhold the combination of up to 20 pairs of transitions and generate 5000 training examples with lengths 3 to 15 and corresponding test data as described in \cref{sec:UC}. 
For iteration generalization, we generate training examples with a maximum iteration count of 3 and test on longer examples of length up to 30 with an iteration count of at least 4.
Since the out-of-distribution performance of two checkpoints of the same model can vary significantly, we report averages on the test set of the last 10 epochs.

\begin{table}[t]
\centering

\scalebox{0.98}{
\begin{tabular}{lrrrr}
\toprule
& \multicolumn{2}{c}{Iteration} & \multicolumn{2}{c}{UC} \\
  & Acc$\uparrow$ & ED$\downarrow$ & Acc$\uparrow$ & ED$\downarrow$ \\
\midrule
ByT5 & 37.8 & 5.87 & 47.4/57.5 & 1.49/0.93 \\
Naive & 42.6 & 4.41 & 44.9/43.2 & 1.52/1.35 \\
Set & 44.4 & 4.58 & 43.6/42.0 & 1.47/1.31 \\
TE & 61.3 & 2.49 & 57.3/63.1 & 1.13/0.74 \\[0.7ex]
\dfourmethod & \textbf{94.8} & \textbf{0.12} & \textbf{73.1/93.3} & \textbf{0.61/0.13} \\
$\ $\added{-prefix} & 84.9 & 0.62 &  	61.1/76.3 & 0.99/0.50 \\ 	
\bottomrule
\end{tabular}}
   \caption{Evaluating systematic generalization on FST tasks with 4 states.  We report averages over 5 tasks. ED is edit distance. Due to an outlier task on UC, we additionally report the median after `/'.}
    \label{tab:within-pretrain}
    \vspace{-12pt}
\end{table}

\paragraph{Results} The results can be found in \cref{tab:within-pretrain}.
On average, \dfourmethod achieves close to perfect accuracy (with one outlier on UC, skewing the mean). TE also shows a clear improvement over the other baselines but \dfourmethod outperforms TE by a large margin. This suggests that \dfourmethod and TE, to a lesser extent, indeed have acquired a stronger inductive bias for FSTs than the other methods. \added{Using \dfourmethod without the tunable prefix leads to a substantial drop in accuracy, highlighting its importance. We analyze the representations learned by \dfourmethod in the tunable prefix in \cref{appendix:analysis-prefix}.}

\subsection{More Complex FSTs}
\label{sec:more-complex}

\begin{figure*}
    \centering
    \includegraphics[height=5.1cm]{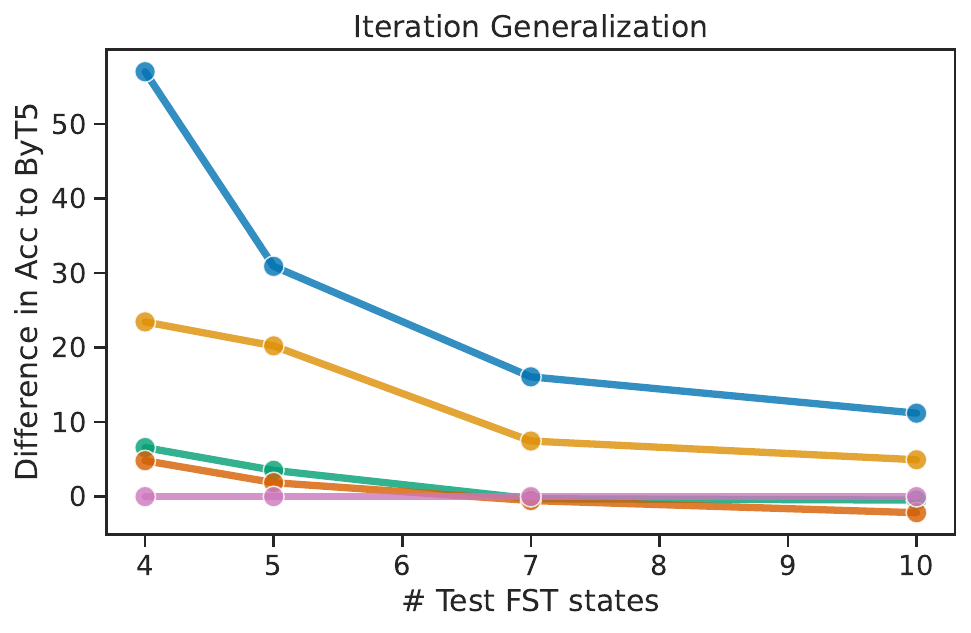}%
    \hfill
    \includegraphics[height=5.1cm]{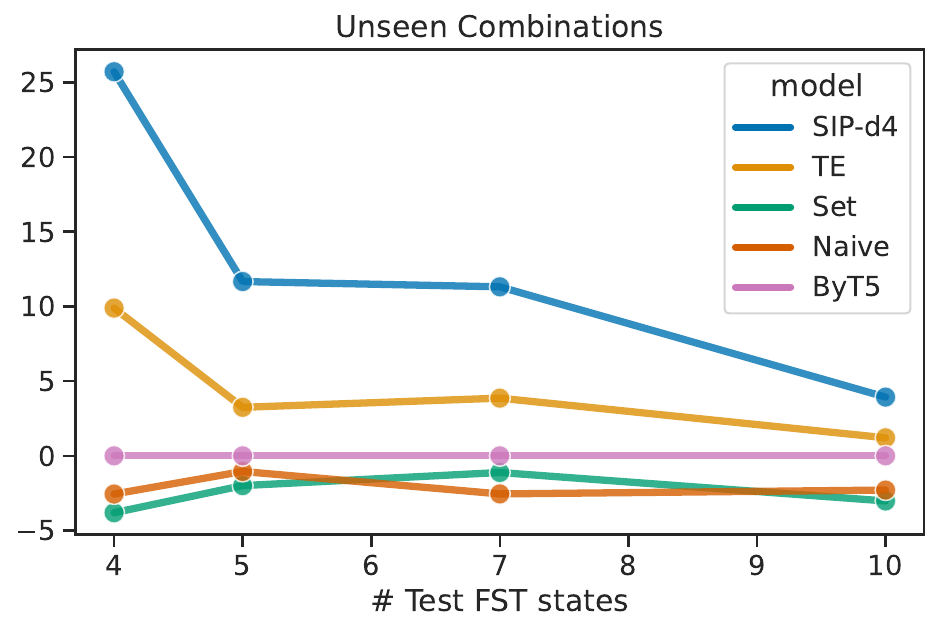}
    \caption{Evaluation on deterministic FST tasks with more states than seen in pre-training. We show the deviation in percentage points from ByT5.}
    \label{fig:more-states-diff-acc}
    \vspace{-12pt}
\end{figure*}

Does the inductive bias introduced by \method extend beyond the pre-training distribution to more complex FST tasks? To investigate this, we use the same sampling methodology but generate FSTs with more states. \dfourmethod was pre-trained on FSTs with up to 4 states, and we evaluate on FST tasks with 5, 7 and 10 states. 

We show in \cref{fig:more-states-diff-acc} how the individual models deviate from the accuracy of ByT5 as a function of the number of states in the test FST. \method always performs best by a clear margin regardless of the number of states in the FSTs. As we increase the number of states and move away from the pre-training distribution, \method improves less over the baselines. We see a similar pattern for TE but with considerably smaller improvements over ByT5.

\subsection{Non-Deterministic FSTs}
\label{sec:non-deterministic}
As shown in the previous section, \method still works well for more complex FST tasks than seen during pre-training. However, this evaluation focused on the favourable case where both pre-training and evaluation involve the same class of FSTs, namely deterministic FSTs. Deterministic FSTs can only take left context into account (see \cref{sec:background}), which is a restrictive assumption.
Here, we evaluate if the inductive bias conferred by \method carries over to non-deterministic functional FSTs, \ie those that can also take context to the \textit{right} into account. 

\begin{table}[t]
\centering

\scalebox{0.97}{
\begin{tabular}{lrrrr}
\toprule
 & \multicolumn{2}{c}{Iteration} & \multicolumn{2}{c}{UC} \\
 & Acc$\uparrow$ & ED$\downarrow$ & Acc$\uparrow$ & ED$\downarrow$ \\
\midrule
ByT5 & 83.4 & 0.52 & 83.1 & 0.40 \\
Naive & 83.1 & 0.49 & 84.2 & 0.37 \\
Set & 82.3 & 0.52 & 83.7 & 0.37 \\
TE & 84.2 & 0.49 & 82.7 & 0.42 \\ [0.7ex]
\dfourmethod & 87.8 & 0.32 & 90.0 & 0.24 \\
\method-d4+ & 88.2 & 0.30 & 90.5 & 0.22 \\
\method-nd7 & \textbf{89.5} & \textbf{0.27} & \textbf{91.2} &\textbf{0.18} \\
\bottomrule
\end{tabular}}
    \caption{Evaluation on non-deterministic FSTs. We report averages over 5 tasks.}
    \label{tab:bimachines}
    \vspace{-12pt}
\end{table}

We automatically generate 5 non-deterministic FSTs with 21 states (see \cref{appendix:bimachines} for details) and report averages in \cref{tab:bimachines}. Despite the structural mismatch between pre-training and the downstream tasks, \dfourmethod shows clear improvements over the baselines. Interestingly, TE does not consistently outperform the other baselines, despite its stronger results on deterministic FSTs.

Our pre-training procedure does not hinge on using deterministic FSTs. This raises the question if we can achieve even better performance by adjusting the inductive bias. To investigate this, we further pre-train \dfourmethod on 40,000 non-deterministic FSTs with up to 7 states, which we call \method-nd7. 
To control for the additional training data of \method-nd7, we also further pre-train \dfourmethod with the same number of deterministic FSTs with the same characteristics as in \cref{sec:construct-pretrain} (\method-d4+). The results in \cref{tab:bimachines} show better performance of \method-nd7, which supports the hypothesis that the inductive bias can be adjusted. \method-d4+ shows a smaller improvement over \dfourmethod. Based on 5 additional FSTs per setup to gain more statistical power, we found that the difference between \method-nd7 and \method-d4+ is statistically significant ($p \approx 0.017$, $n=20$, paired approx. permutation test).

\section{Transfer to Natural Data}
In this section, we investigate to what degree the inductive bias from pre-training on synthetic data transfers to tasks with natural data that have been traditionally approached with finite state methods.

\subsection{Low-resource Grapheme-to-Phoneme Conversion}
\label{sec:g2p}
Grapheme-to-phoneme conversion is the task of converting a word as a sequence of symbols (for example, letters in the Latin alphabet) into a description of how this word is pronounced as letters in the IPA alphabet. For example, a possible pronunciation of `explanation' is \textipa{[""Ekspl@"neIS@n]}. Grapheme-to-phoneme conversion can be part of text-to-speech pipelines and FSTs for this purpose usually are two or three magnitudes larger than the FSTs we constructed for pre-training. Because of this, it enables us to test how far beyond the pre-training distribution \method remains helpful. 
We focus on learning from small amounts of data, for which a structural inductive bias towards FSTs should be particularly helpful. We evaluate on 7 low-resource languages from different language families that use their own scripts (Balinese, Coptic, Gothic, Lao, Sylheti, Telugu and Central Atlas Tamazight). We obtained the data from Wikipron \citep{lee-etal-2020-massively}.

\begin{table}[t]
    \centering

\resizebox{\linewidth}{!}{
\addtolength{\tabcolsep}{-0.25em}
\begin{tabular}{lrrrrrrr|r}
\toprule
& ban & cop & got & lao & syl & tel & tzm & Avg \\
\midrule
\grayout{Charsiu} & \grayout{68.3} & \grayout{7.8} & \grayout{67.0} & \grayout{35.1} & \grayout{47.6} & \grayout{73.3} & \grayout{18.6} & \grayout{45.4} \\
ByT5 & 50.2 & 1.0 & 30.7 & 1.9 & 9.8 & 6.9 & 2.7 & 14.8 \\
Set & 53.9 & 2.2 & 58.2 & 5.8 & 28.2 & 27.7 & 6.4 & 26.1 \\
TE & 54.7 & 1.9 & 37.0 & 5.1 & 30.0 & 16.2 & 7.4 & 21.8 \\
\dfourmethod & \textbf{59.2} & \textbf{6.6} & 56.5 & \textbf{8.2} & \textbf{39.8} & \textbf{33.1} & \textbf{11.0} & \textbf{30.6} \\
$\ $ -prefix & 55.1 & 3.2 & \textbf{63.9} & 7.8 & 28.0 & 28.9 & 7.0 & 27.7 \\
\bottomrule
\end{tabular}
\addtolength{\tabcolsep}{0.25em}
}
\caption{Grapheme-to-phoneme conversion with 100 training examples. We show averages of 5 selections of training examples.}
    \label{tab:g2p}
\vspace{-12pt}
\end{table}

As a soft upper bound, we compare with Charsiu \citep{zhu22-interspeech} which is a ByT5-small model that has been further pre-trained on 7.2 million examples of grapheme-to-phoneme conversion across 100 languages.  Although Charsiu was not exposed to the scripts of the languages we chose, it may have seen related languages whose scripts are encoded similarly in Unicode.

We report accuracies in \cref{tab:g2p}, and phoneme-error-rates in \cref{app:g2p-full}; trends are identical. The original ByT5-small model performs worst on average despite being a strong model for grapheme-to-phoneme conversion in general \citep{xue2022byt5}.
On average across the languages, \dfourmethod outperforms the other methods that pre-train on synthetic data as well as ByT5. The difference between \dfourmethod and Set is statistically significant ($p \approx 4 \times 10^{-4}$, paired approx.\ permutation test).
\added{Fine-tuning \dfourmethod without the tunable prefix leads to a drop in performance, except for Gothic.}
Charsiu performs very well on Telugu, potentially because of its large overlap in lexicon with Sanskrit \citep{Sanskritization}, which is part of its training data.

\subsection{Few-shot text editing}
Learning simple text editing tasks (Jane Doe $\rightarrow$ J. Doe) from a handful of examples with a Transformer requires a strong structural inductive bias to overcome competing explanations of the data and hence provides a good benchmark for our approach. While current LLMs may seem like the ideal choice for such tasks, they are prone to hallucinations, \eg ignoring the input and resorting to frequent entities (see \cref{appendix:hallucination} for an example). 

Text editing has been studied in the context of program synthesis and we evaluate on 19 such tasks from the SyGuS competition 2017 \citep{alur2017sygus}. Instead of predicting a program, our model directly operates on input/output examples. 
We note that 17 of these tasks can be solved by compact FSTs, whereas two cannot. These two tasks are \textit{rev-name} (Jane Doe $\rightarrow$ Doe Jane) and \textit{sur-initial} (John Doe $\rightarrow$ Doe, J.), which require tracking information about the first name in the states.

\begin{table}[t]
    \centering

\addtolength{\tabcolsep}{-0.38em}

\resizebox{\linewidth}{!}{
\begin{tabular}{lrrrrrr|rr}
\toprule
 & \multicolumn{2}{c}{rev-name} & \multicolumn{2}{c}{sur-initial} & \multicolumn{2}{c}{FST} & \multicolumn{2}{c}{Overall} \\
 & Acc$\uparrow$ & ED$\downarrow$ & Acc$\uparrow$ & ED$\downarrow$ & Acc$\uparrow$ & ED$\downarrow$ & Acc$\uparrow$ & ED$\downarrow$ \\
\midrule
ByT5 & 11.8 & 6.81 & 47.2 & 1.76 & 47.6 & 1.42 & 45.7 & 1.72 \\
Charsiu & 43.8 & 1.73 & 52.8 & 0.87 & 62.4 & 0.74 & 60.9 & 0.80 \\
Set & 79.0 & 1.34 & 41.5 & 3.37 & 68.2 & 0.71 & 67.4 & 0.89 \\
TE & 80.3 & 1.08 & 88.2 & 0.41 & \textbf{95.7} & \textbf{0.11} & \textbf{94.5} & 0.17 \\
\dfourmethod & 92.4 & 0.34 & \textbf{97.2} & \textbf{0.10} & 91.6 & 0.13 & 91.9 & \textbf{0.14} \\
$\ $\added{-prefix} & \textbf{97.8} & \textbf{0.10} & 72.6 & 0.51 & 89.0 & 0.27 & 91.4 & 0.18 \\
\bottomrule
\end{tabular}}
    \caption{Averages of accuracy and edit distance across 5-shot text editing tasks based on 8 draws of training examples. We report results grouped by tasks that cannot be solved by a compact FST (rev-name, sur-initial), tasks that can be solved by FSTs, and overall averages.}
    \label{tab:text-editing}
\vspace{-12pt}

\addtolength{\tabcolsep}{0.38em}
\end{table}

We report results for 5-shot experiments in \cref{tab:text-editing}.
\dfourmethod and TE excel at this, reaching well above 90\% accuracy on average whereas the other methods perform worse by a large margin. Charsiu does not perform clearly better than baselines such as Set -- even though it obtains excellent results on grapheme-to-phoneme conversion. 
Interestingly, TE performs better than \dfourmethod on the tasks that can be solved with FSTs, potentially because the initialization of the prefix for TE follows the same distribution as during pre-training, which is not the case for \method.
However, \method considerably outperforms TE on the two tasks that cannot be compactly represented by FSTs, suggesting that some of the dynamics acquired during pre-training can sometimes be leveraged in other contexts as well. \added{Fine-tuning \dfourmethod without the tunable prefix leads only to a very small drop in accuracy on average.}

\section{Analysis: \method leads to FST simulation}
\label{sec:analysis}
We motivated our approach by the hypothesis that \method's pre-training encourages the model to simulate FSTs internally, and that this provides the structural inductive bias. In this section, we present evidence that (i) SIP models indeed approximately simulate FSTs in the hidden states, and (ii) that the dynamics responsible for simulation are re-used after fined-tuning all parameters on input/output pairs only.

\begin{figure}[t]
    \centering
    \includegraphics[width=0.97\linewidth]{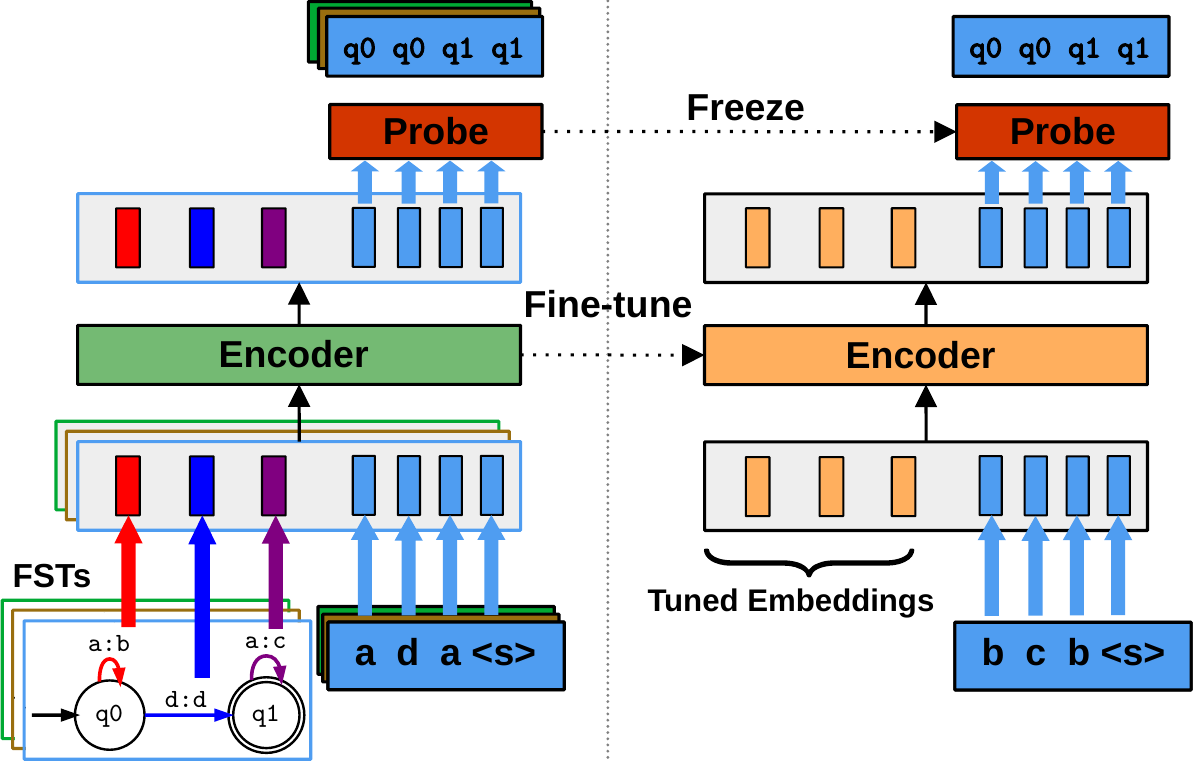}
    \caption{Left: we train a linear probe on the encoder representations of a \method pre-trained model to predict for each input token $x_i$ which state the encoded FST is in before processing $x_i$. The end-of-sequence token is represented as \texttt{<s>}. Right: we freeze the trained probe, fine-tune the \method model on input/output pairs and extract state sequences from it with the probe.}
    \label{fig:probe}
    \vspace{-12pt}
\end{figure}

For a model to simulate FSTs in its hidden representations, it must be able to track the FST state when processing a string, and it should be possible to extract the FST state with a probe. To test this, we mirror the pre-training setup and provide \dfourmethod with an FST and an input string (\cref{fig:probe}, left). For each token, we extract the top-layer activations of the encoder, and learn a linear probe with a softmax layer to predict the ID of the state that the given FST is in \textit{before} processing that token. Since state IDs are largely arbitrary, the probe has to learn to relate the hidden representations to the FST presented in the input.

The probe achieves 99.3\% token-level accuracy on a test set with unseen FSTs, and a whole-sequence accuracy of 93.9\%. We also evaluate a trivial heuristic that returns a random state that has an appropriate outgoing transition for each token in the input. This heuristic achieves a token-level accuracy of 68.9\%, and a whole-sequence accuracy of only 17.8\%. 
A probe trained on ByT5 representations, \ie before \method pre-training, performs even worse at 42.9\% token-level accuracy and whole-sequence accuracy of only 7.1\% (see \cref{appendix:probe-non-sip}).
Hence, the model has learned a non-trivial way to simulate transitions between states of the FST encoded in the prefix. This is remarkable because the pre-training procedure for \dfourmethod does not provide supervision for \textit{how} to process strings.

While this shows that \method leads to the simulation of state transitions after pre-training, does the model leverage the simulation ability on downstream tasks? Recall that we fine-tune all parameters of the model (\cref{sec:fine-tune}), so the model could employ a very different strategy to fit the data.
To investigate this, we set the trained probe aside and freeze it (\cref{fig:probe}, right). We then fine-tune \dfourmethod on the iteration generalization tasks with 4 states in the gold FST (cf. \cref{tab:within-pretrain}). Finally, we apply the frozen probe to the fine-tuned model to see if the state sequences we extract are similar to those of the ground truth FST. Fine-tuning \dfourmethod could induce the same FST as the ground truth but use a different numbering of the states. To account for this, for each of the five tasks, we find the isomorphism between the predicted state IDs and the ground truth that gives the best match on average.

\begin{figure}
    \centering
    \includegraphics[width=0.495\linewidth]{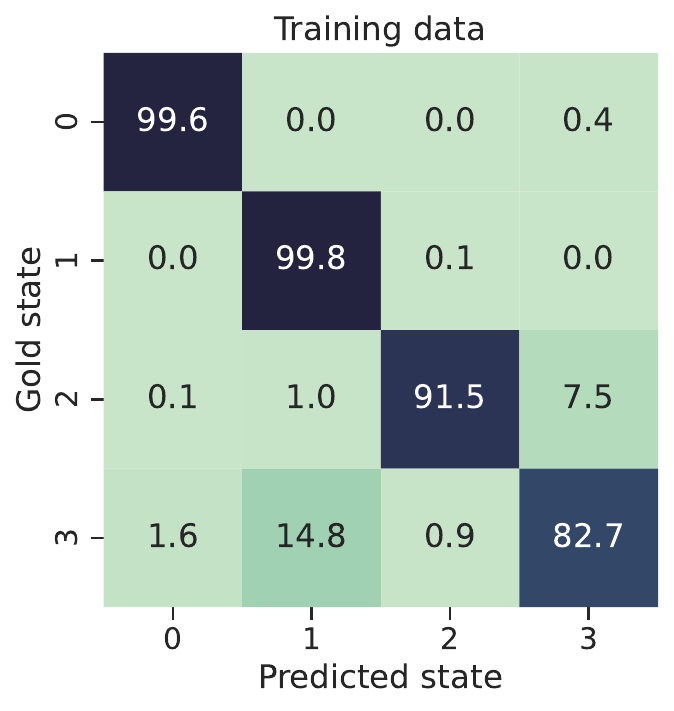}%
    \hfill
    \includegraphics[width=0.495\linewidth]{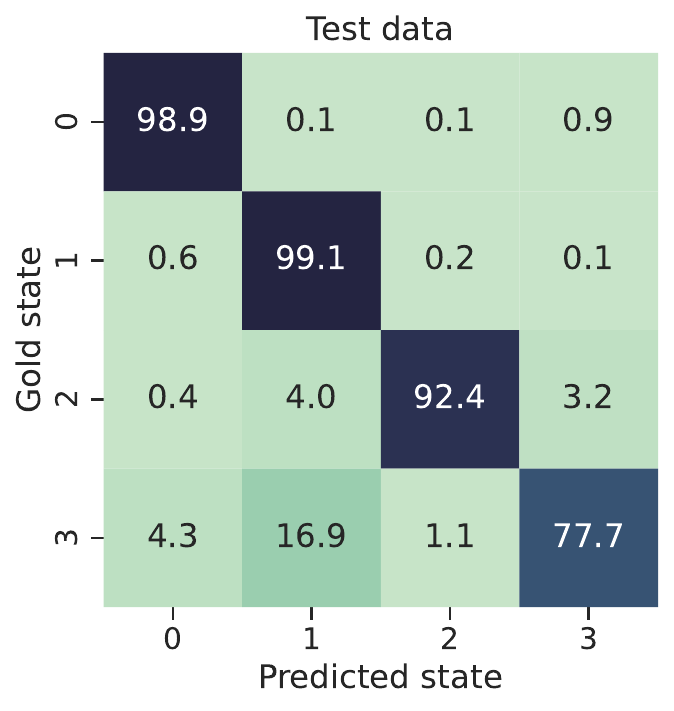}
    \caption{Row-normalized confusion matrices on the training and test data between ground truth and the state predicted by the frozen probe applied to fine-tuned models. We average across the 5 iteration generalization tasks (\cref{sec:within-pretrain}). 
    }
    \label{fig:confusion-matrices}
    \vspace{-12pt}
\end{figure}

The results are presented in \cref{fig:confusion-matrices} as confusion matrices between predicted and gold states. The probe extracts state sequences that resemble the state sequences of the gold FST (up to isomorphism), both on the training data and the out-of-distribution test data. 
We also find that deviation from the ground truth state sequence correlates with errors by the fine-tuned model: if the probe extracts correct state sequences, the model achieves an accuracy of 98.6\% on the iteration generalization tasks, whereas it drops to 89.8\% when the probe extracts state sequences that deviate. The difference is statistically significant (approximate permutation test, $p \approx 5 \times 10^{-5}$).
Overall, this shows that the fine-tuned model reuses the dynamics for state tracking and learns representations similar to the ground truth FST.

\section{Conclusion}

We present \method, a simple, efficient and adjustable method for introducing a structural inductive bias into a seq2seq model. We focus on an inductive bias towards FSTs, one of the simplest computational devices that is useful for NLP applications. We achieve this by pre-training a Transformer to simulate automatically generated FSTs, \ie to predict the output of an FST given an input string and a description of the FST. 
Our experiments show that our method imparts the desired inductive bias, resulting in improved systematic generalization and better few-shot learning for FST-like tasks. In addition, we show with probing experiments that a model trained with \method simulates transitions between FST states in its hidden representations, and that the dynamics behind this are leveraged during fine-tuning.
In future work, we plan to extend this methodology to more expressive formalisms such as Pushdown Transducers which can be used for a wider range of downstream NLP tasks.

\section*{Limitations}

Our investigation focuses on FSTs with a relatively small number of states. However, the results in \cref{sec:more-complex} and in the experiments on grapheme-to-phoneme conversion show that even pre-training with FSTs with a small number of states has positive impacts for tasks that require larger or more complex FSTs.

The probing experiments show that the model simulates transitions between states similarly to an FST, but we did not perform a mechanistic interpretation of how exactly this is implemented in the weights. One potential mechanism behind the simulation behaviour is the construction of \citet{liu2022transformers} who show that Transformer decoders can simulate transitions between states of deterministic finite automata for strings of length up to $n$ using $O(\log(n))$ layers.

Acquiring a specific inductive bias by means of learning to simulate a computational device is a general idea that could be applicable beyond FSTs but might be unsuitable in cases where (i) it is difficult to formulate a reasonable computational device to simulate (such as document classification and sentiment analysis beyond keyword spotting), or (ii) the computational device would be very hard or infeasible to simulate (\eg Turing machines).

Our experiments focus on moderately sized models (300M parameters) with an encoder-decoder architecture, and we did not investigate large decoder-only models. Our methodology can also be applied to decoder-only models, and we do not foresee any reasons why it could be less effective in that setup.

Finally, we only consider the standard Transformer architecture and we leave it to future work to explore the impact of \method on variants of the Transformer architecture designed for handling long character sequences \citep{yu2023megabyte} or in the context of state-space models \citep{wang2024mambabyte}.

\ifreviewmode
\else

\section*{Acknowledgements}
We thank Verna Dankers, Victor Prokhorov, and Christine Sch\"afer for discussions and comments.
ML is supported by the UKRI Centre for Doctoral Training in Natural Language Processing, funded by the UKRI (grant EP/S022481/1), the University of Edinburgh, School of Informatics and School of Philosophy, Psychology \& Language Sciences, and a grant from Huawei Technologies. IT is supported by the Dutch National Science Foundation (NWO Vici VI.C.212.053).

\fi

\renewcommand{\baselinestretch}{1.0}

\bibliography{bib}

\appendix

\section{Generation of Synthetic Data and Splits}

\subsection{Generating deterministic FSTs}
\label{appendix:deterministic-fsts}

Before describing our procedure for sampling deterministic FSTs, we briefly establish notation. An FST is a tuple $\langle Q, \Sigma, \Gamma, I, F, \Delta \rangle$, where $Q$ is a finite set of states, $\Sigma$ is the input alphabet, $\Gamma$ is the output alphabet, $I \subseteq Q$ is a set of initial states, $F \subseteq Q$ is a set of final states and $\Delta \subseteq Q \times (\Sigma \cup \{ \epsilon\}) \times (\Gamma \cup \{ \epsilon\}) \times Q$ are the transitions. We assume $\Sigma=\Gamma$ and call it $V$ for vocabulary. 

For technical reasons, we exclude the three characters \texttt{[}, \texttt{]} and \texttt{\textbackslash} from the vocabulary as they are interpreted as special characters by OpenFST, which we use for constructing and representing FSTs.

In addition to the shorthand for identity transitions (\texttt{id}), we also have shorthands for converting upper case to lower case and vice-versa (\texttt{lower-to-upper}, \texttt{upper-to-lower}). We describe our procedure to generate a deterministic FST with pseudocode in \cref{alg:det-fst}. It receives as argument $n$ (the number of states in the FST), $f$ (number of final states), $V$ (the vocabulary of this FST), and probabilities \textsc{p-id}, \textsc{p-drop}, \textsc{p-shorthand}. These probabilities control the likelihood of using a shorthand, not drawing an outgoing edge (\textsc{p-drop}) with a given symbol, and creating a single identity transition (\textsc{p-id}).
We use \textsc{choice} to denote a uniform random choice from a finite set.

We use $\textsc{p-id}=0.2, \textsc{p-drop}=0.4, \textsc{p-shorthand}=0.15$ in our experiments.

\algrenewcommand\algorithmicif{\textbf{with prob}}
\algrenewcommand\algorithmicthen{}

\algrenewcommand\algorithmicindent{0.51em}%

\begin{algorithm}[h]
	\caption{Generate a random deterministic FST}
	
	\begin{algorithmic}
		\Function{gen-det-fst}{$n, f, V, \textsc{p-id}, \textsc{p-drop}, $ $ \textsc{p-shorthand}$}
		\State $Q = \{0, \ldots n-1\}$
            \State $\Delta = \emptyset$
            \State $I = \{0\}$
		\For{$q \in Q$}
                \State $q' = \textsc{choice}(Q)$
    		\If{\textsc{p-shorthand}}
                    \State $s = \textsc{choice}([\texttt{id}, $
                    \State $\quad \texttt{lower-to-upper}, \texttt{upper-to-lower}])$
                    \State $\Delta := \Delta  \cup \{ q \trans{s}{s} q') \}$
                \Else
                    \For{$\sigma \in V$}
                     \If{\textsc{p-drop}}
                     \State no-op
                     \Comment{No outgoing edge with $\sigma$ at $q$}
                     \ElsIf{\textsc{p-id}}
                     \State $\Delta := \Delta  \cup \{ q \trans{\sigma}{\sigma} q' \}$
                     \Else 
                     \State $\Delta := \Delta  \cup \{ q \trans{\sigma}{\textsc{choice}(V \cup \{\epsilon\})} q' \}$
                     \EndIf
                    \EndFor
                \EndIf
		\EndFor
            \State Eliminate states from $Q$ through which no accepting path can go
            \State Choose random subset $F$ of $Q$ with $|F| = \min(f, |Q|)$
		\State \textbf{return} minimized FST with states $Q$, transitions $\Delta$, initial states $I$ and final states $F$
		\EndFunction
	\end{algorithmic}
	\label{alg:det-fst}
\end{algorithm}

\subsection{Generating Non-deterministic Functional FSTs}
\label{appendix:bimachines}
\begin{figure}[t]
    \centering
    \includegraphics[width=\linewidth]{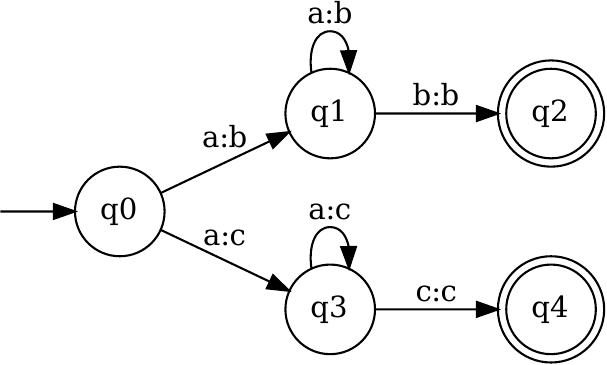}
    \caption{A functional but non-deterministic FST.}
    \label{fig:fst-from-bimachine}
\end{figure}

\begin{figure}[t]
    \centering
    \includegraphics[width=\linewidth]{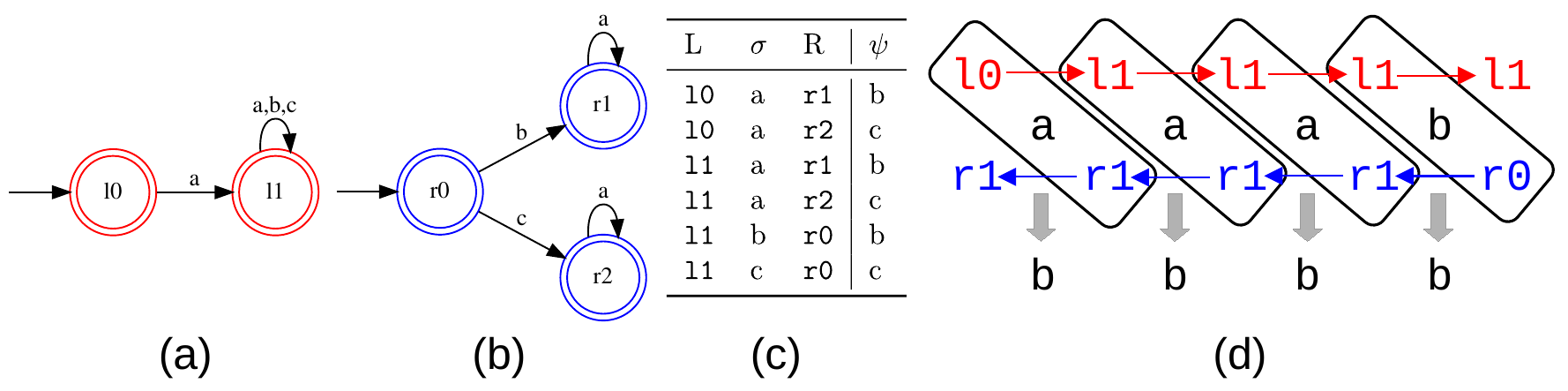}
    \caption{(a) - (c) shows a bimachine that is equivalent to \cref{fig:fst-from-bimachine}. (a) Left automaton $A^l$, (b) Right automaton $A^r$, (c) output function $\psi$. (d) shows an example run of the bimachine on the input \texttt{aaab} which is mapped to \texttt{bbbb}.}
    \label{fig:bimachine}
\end{figure}

\begin{algorithm}[h]
	\caption{Generate output function for bimachine}
	
	\begin{algorithmic}
		\Function{gen-output-$\psi$}{$n^L, n^R, V, \textsc{p-id} = 0.2$}
		\For{$q^L \in 0, \ldots, n^L-1$}
                \For{$q^R \in 0, \ldots, n^R-1$}
                    \For{$\sigma \in V$}
                        \If{\textsc{p-id}}
                            \State $\psi(q^L, \sigma, q^R) := \sigma$
                        \Else
                            \State $\psi(q^L, \sigma, q^R) := \textsc{choice}(V \cup \{ \epsilon\})$
                        \EndIf
                    \EndFor
                \EndFor
                
        	\EndFor
		\State \textbf{return} $\psi$
		\EndFunction
	\end{algorithmic}
	\label{alg:gen-psi}
\end{algorithm}

It is not straightforward to directly generate non-deterministic FSTs that are guaranteed to express a function. However, we can directly generate a bimachine, which then can be converted into an FST.

Bimachines \citep{schutzbnbergee1961remark} represent the functions expressible by FSTs, \ie for every functional FST there is a bimachine that represents it (and vice-versa). A bimachine consists of two deterministic finite state automata (called left and right) and an output function. Let $A^L$ be the left FSA with states $Q^L$ and transition function $\delta^L: Q^L \times \Sigma \rightarrow Q^L$), and let $A^R$ bet the right FS with states $Q^R$ and transition function $\delta^R: Q^R \times \Sigma \rightarrow Q^R$. 
The output function is $\psi: Q^l \times \Sigma \times Q^r \rightarrow \Gamma^*$. 
All states of $A^L$ and $A^R$ are final states. Given an input string $x=\sigma_1\sigma_2\sigma_3\ldots\sigma_n$, a bimachine runs $A^L$ from left to right over $x$, keeping track of the states $q^l_0, q^l_1, q^l_2, \ldots q^l_n$. It also runs $A^R$ over the string $x$ but this time from right to left, again keeping track of the states $q^r_0, q^r_1, q^r_2, \ldots q^r_n$ that are visited. Then, the state sequence of the right automaton is reversed and $\psi$ is applied `elementwise' as illustrated in \cref{fig:bimachine}. More formally, the output of the bimachine is $\psi(q^l_0, \sigma_1, q^r_{n-1}) \psi(q^l_1, \sigma_1, q^r_{n-2}) \ldots \psi(q^l_{n-1}, \sigma_1, q^r_0)$.

Bimachines can be compiled into FSTs with a simple product construction. For a bimachine $\langle A^L, A^R, \psi \rangle$, one can construct an equivalent FST as follows: $$\langle Q^L \times Q^R, \Sigma, \Gamma, \{s^L\} \times Q^R, Q^L \times \{s^R\}, \Delta \rangle $$
where $s^L$ and $s^R$ are initial states of $A^L$ and $A^R$, and $\Delta$ contains all transitions 
\begin{align*}
    \Delta = \{ \langle q^L, q^R \rangle \trans{\sigma}{\rho} \langle q'^L, q'^R \rangle \  | \  & \delta^L(q^L, \sigma) = q'^L, \\
    &\delta^R(q'^R, \sigma) = q^R, \\
    & \rho = \psi(q^L, \sigma, q'^R) \}
\end{align*}
We refer to \citet{mihov-schulz-2019} for details and further information about bimachines.

To sample bimachines, we re-use \cref{alg:det-fst} with $\textsc{p-shorthand}=0$, and ignore the outputs of the transitions, treating them as FSAs. We sample the output function according to \cref{alg:gen-psi}. For the test data creation (\cref{tab:bimachines}), we use 5 states in the left FSA and 4 states in the right FSA, and set $\textsc{p-drop}=0.4$. For creating the training data for \method-nd7, we use 2 or 3 states in either left or right automaton and set $\textsc{p-drop}=0.6$ to keep the length of the prefix low to save GPU memory.

\subsection{Unseen Combinations of Transitions}
\label{appendix:uc}

We now describe the construction by which we create training and test data for the evaluation of Unseen Combinations of transitions. We first describe how we construct an FST for the training and test data, respectively, given a choice of transitions whose combination we want to withhold. Then, we briefly describe how those transitions are chosen.

\begin{figure}[t]
    \centering
    \includegraphics[width=\linewidth]{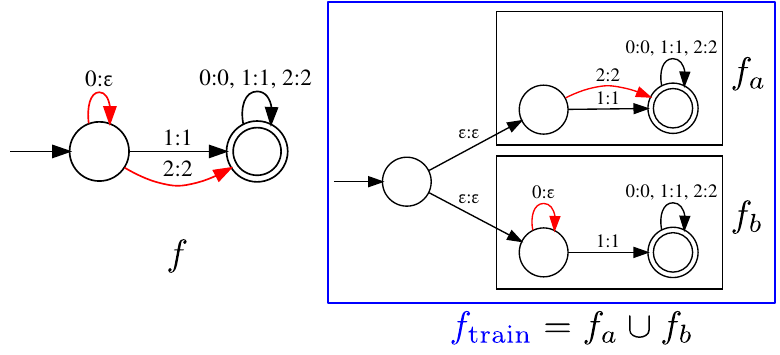}
    \caption{Constructing training data for evaluating unseen combinations of transitions. Based on the given FST $f$, we construct an FST $f_{\text{train}}$ that withholds the \textit{combination} of the two red transitions.}
    \label{fig:edge-deletion}
    \vspace{-4pt}
\end{figure}

Given an FST $f$ (illustration in \cref{fig:edge-deletion}, left) and transitions $t_a$ and $t_b$ (highlighted in red) whose combination we want to withhold, we construct a new FST $f_{\text{train}}$ as follows: We create two copies $f_a, f_b$ of the original FST $f$. In $f_a$, we remove the transition $t_b$; in $f_b$, we remove the transition $t_a$. Then $f_{\text{train}} = f_a \cup f_b$, which can be constructed by introducing a new initial state with $\epsilon$-transitions into the respective initial states of $f_a$ and $f_b$ (right side of \cref{fig:edge-deletion}). 
This ensures that any accepting path goes through $f_a$ or $f_b$ but cannot alternate between the two. 
Hence, $t_a$ or $t_b$ can be used -- but not both in the same string. 
Note that $f_{\text{train}}$ still describes a partial function (rather than a relation) because any accepting path in $f_a$ and any accepting path in $f_b$ is also an accepting path in $f$. As a result, whenever $f_a$ and $f_b$ are both defined, they agree on the result $f_a(x) = f_b(x) = f(x)$. We test exclusively for how a model handles unseen combinations of transitions by generating examples from $f$ for which $f_{\text{train}}$ is \textit{not} defined.

To make the generalization setup more challenging, these steps can be applied to multiple pairs of adjacent transitions at the same time, \ie to withhold $\langle t^1_{a}, t^1_{b} \rangle, \ldots, \langle t^k_a, t^k_b \rangle$: We create the copy $f_a$ and remove the transitions $t^1_b, \ldots, t^k_b$ from $f_a$ and analogously remove $t^1_a, \ldots, t^k_a$ from $f_b$.

Now, we briefly describe how we select \textit{which} pairs of transitions we want to withhold. We only select adjacent transitions, \ie transitions where one can be used immediately after the other, excluding self-loops. In addition, some transitions cannot be deleted without cutting off a vital initial or final state, which can lead to $f_{\text{train}}$ being undefined for any string (and hence no training data). We ensure this never happens by never withholding the first transition into each state based on a depth-first traversal of the FST.

\subsection{Additional dataset information}
For all experiments with synthetic data (generated by FSTs), we generate 5000 training examples and 1000 test examples. To reduce variance across tasks, we fix the vocabulary size to its maximum value (25) in the pre-training data and choose the vocabulary only from the printable ASCII characters.

\paragraph{Length distribution} The input strings in the pre-training data we generate for \dfourmethod have a minimum length of 1, an average length of 15.57 and a maximum length of 35. We report the length distributions for the iteration generalization experiments in \cref{sec:evaluating-inductive-bias} in \cref{tab:length-dist-eval}. 

\begin{table}[]
    \centering
\begin{tabular}{rlrrr}
\toprule
 Num. states & Split & Min & Max & Mean \\
\midrule
 4 & train & 2 & 11 & 4.66 \\
 4 & test & 4 & 30 & 18.97 \\
 5 & train & 2 & 14 & 5.39\\
 5 & test & 4 & 30 & 19.53 \\
 7 & train & 2 & 20 & 6.12 \\
 7 & test & 4 & 30 & 20.13 \\
 10 & train & 2 & 25 & 7.31 \\
 10 & test & 4 & 30 & 20.62 \\ [0.5ex]
 21 & train & 2 & 30 & 11.80 \\
 21 & test & 5 & 30 & 23.07\\
\bottomrule
\end{tabular}
    \caption{Distribution of input lengths of the train/test data we generate for the iteration generalization experiments in \cref{sec:evaluating-inductive-bias}. The tasks with 21 states are the non-deterministic FSTs from \cref{sec:non-deterministic}.}
    \label{tab:length-dist-eval}
\end{table}

\paragraph{SyGuS} We took the data from the SyGuS competition github \url{https://github.com/SyGuS-Org/benchmarks/tree/master/comp/2017/PBE\_Strings\_Track}, and extracted the `constraints'. For each text editing tasks, there are usually three files, e.g. \texttt{firstname}, \texttt{firstname-long}, \texttt{firstname-long-repeat}. We only consider data from the \texttt{*-long} variant because the non-marked variant (e.g. \texttt{firstname}) is a subset of the \texttt{*-long} variant, and we exclude \texttt{*-long-repeat} as it contains repeated data points. We also exclude some text editing tasks that have insufficient amounts of data for reliable evaluation (\texttt{bikes}) and some tasks where the input is not a single string but a pair of strings if concatenating the strings results in particularly long inputs (\texttt{univ}), or if the concatenation of the string pair makes the task trivial (name-combine, which would correspond to an identity operation). For a few-shot experiment, we sample 5 training examples and evaluate on the rest.

We note that the original intention in the design of the benchmark data was for program synthesis rather than few-shot learning. The data contains names, and separately it contains phone numbers (but not combined). However, we believe both to be synthetically generated.

\paragraph{Grapheme-to-phoneme} We obtain data from \citet{lee-etal-2020-massively}, and conduct experiments mainly on the broad transcription, except for Telugu and Tamazight, where we use the narrow transcription. For each experiment, we randomly sample 100 training examples, and use the rest as test data. The data is available under a permissible license: \url{https://en.wiktionary.org/wiki/Wiktionary:Copyrights}

\section{Additional Results}

\subsection{Additional Results with More States}
\label{appendix:additional-results-more-states}
In \cref{fig:more-states-diff-acc}, we show accuracy relative to the accuracy of ByT5. Here, we show the absolute accuracies and edit distances in \cref{tab:more-states-full}.

\begin{table*}[t]
\centering
\begin{tabular}{ll|rr|rr|rr|rr}
\toprule
 & Num States & \multicolumn{2}{c}{4} & \multicolumn{2}{c}{5} & \multicolumn{2}{c}{7} & \multicolumn{2}{c}{10} \\
Gen. Type & Model & Acc$\uparrow$ & ED$\downarrow$ & Acc$\uparrow$ & ED$\downarrow$ & Acc$\uparrow$ & ED$\downarrow$ & Acc$\uparrow$ & ED$\downarrow$ \\
\midrule
\multirow[t]{5}{*}{Iteration} & ByT5 & 37.8 & 5.87 & 58.7 & 3.21 & 48.2 & 3.71 & 45.7 & 3.87 \\
 & Naive & 42.6 & 4.41 & 60.5 & 2.20 & 47.7 & 3.16 & 43.6 & 3.65 \\
 & Set & 44.4 & 4.58 & 62.2 & 2.41 & 48.0 & 3.49 & 45.3 & 3.71 \\
 & TE & 61.3 & 2.49 & 78.9 & 0.86 & 55.7 & 2.29 & 50.7 & 2.95 \\
& SIP-d4 & \textbf{94.8} & \textbf{0.12} & \textbf{89.6} & \textbf{0.27} & \textbf{64.3} & \textbf{1.34} & \textbf{56.9} & \textbf{2.39} \\
\cline{1-10} 
\multirow[t]{5}{*}{UC} & ByT5 & 47.4 & 1.49 & 62.6 & 1.05 & 61.9 & 1.29 & 54.1 & 1.70 \\
 & Naive & 44.9 & 1.52 & 61.6 & 1.08 & 59.3 & 1.30 & 51.8 & 1.68 \\
 & Set & 43.6 & 1.47 & 60.6 & 1.09 & 60.8 & 1.31 & 51.1 & 1.71 \\
 & TE & 57.3 & 1.13 & 65.9 & 0.98 & 65.7 & 1.17 & 55.3 & 1.60 \\
& SIP-d4 & \textbf{73.1} & \textbf{0.61} & \textbf{74.3} & \textbf{0.69} & \textbf{73.2} & \textbf{0.85} & \textbf{58.0} & \textbf{1.44} \\
\bottomrule
\end{tabular}
\caption{Evaluation on deterministic FSTs with more states, showing absolute accuracies and edit distances, corresponding to \cref{fig:more-states-diff-acc}}
\label{tab:more-states-full}
\end{table*}

\subsection{Full results for grapheme-to-phoneme conversion}
\label{app:g2p-full}

\cref{tab:full-g2p} shows the full results of our grapheme-to-phoneme conversion experiments, including phoneme error rate (PER).

\begin{table*}[t]
    \centering

\addtolength{\tabcolsep}{-0.38em}

\begin{tabular}{lrrrrrrrrrrrrrr|rr}

\toprule
 & \multicolumn{2}{c}{ban} & \multicolumn{2}{c}{cop} & \multicolumn{2}{c}{got} & \multicolumn{2}{c}{lao} & \multicolumn{2}{c}{syl} & \multicolumn{2}{c}{tel} & \multicolumn{2}{c}{tzm} & \multicolumn{2}{c}{Avg} \\
 & Acc & PER & Acc & PER & Acc & PER & Acc & PER & Acc & PER & Acc & PER & Acc & PER & Acc$\uparrow$ & PER$\downarrow$ \\
\midrule
\grayout{Charsiu} & \grayout{68.3} & \grayout{.110} & \grayout{7.8} & \grayout{.579} & \grayout{67.0} & \grayout{.067} & \grayout{35.1} & \grayout{.238} & \grayout{47.6} & \grayout{.196} & \grayout{73.3} & \grayout{.070} & \grayout{18.6} & \grayout{.403} & \grayout{45.4} & \grayout{.238} \\
ByT5 & 50.2 & .233 & 1.0 & .847 & 30.7 & .269 & 1.9 & .760 & 9.8 & .598 & 6.9 & .597 & 2.7 & .851 & 14.8 & .594 \\
Set & 53.9 & .216 & 2.2 & .742 & 58.2 & .094 & 5.8 & .595 & 28.2 & .353 & 27.7 & .293 & 6.4 & .658 & 26.1 & .421 \\
TE & 54.7 & .183 & 1.9 & .756 & 37.0 & .174 & 5.1 & .573 & 30.0 & .309 & 16.2 & .377 & 7.4 & .644 & 21.8 & .431 \\
\dfourmethod & \textbf{59.2} & \textbf{.152} & \textbf{6.6} & \textbf{.563} & 56.5 & .096 & \textbf{8.2} & \textbf{.498} & \textbf{39.8} & \textbf{.252} & \textbf{33.1} & \textbf{.228} & \textbf{11.0} & \textbf{.544} & \textbf{30.6} & \textbf{.333} \\
 $\ $ \added{-prefix} & 55.1 & .168 & 3.2 & .681 & \textbf{63.9} & \textbf{.072} & 7.8 & .508 & 28.0 & .333 & 28.9 & .252 & 7.0 & .593 & 27.7 & .372 \\
\bottomrule
\end{tabular}
\caption{Grapheme-to-phoneme conversion with 100 training examples. We show averages of 5 selections of training examples. PER is Phoneme Error Rate: edit distance / length of gold output (lower is better).}
    \label{tab:full-g2p}

\addtolength{\tabcolsep}{0.38em}
\vspace{-6pt}
\end{table*}

\subsection{Additional results with T5-Base}
\label{appendix:t5-base}
We run a subset of the experiments starting off from a pre-trained T5-Base \citep{10.5555/3455716.3455856} instead of ByT5. This model is about one-third smaller than ByT5 (around 200 million instead of 300 million parameters). T5-Base uses a different vocabulary than ByT5, so we resize the output layer to the vocabulary size of ByT5 and re-initialize it. For the input embeddings, we re-purpose the first $n$ embeddings in the T5-Base embedding matrix to represent the token ids according to the ByT5 tokenizer. While this is suitable as a starting point for further pre-training, we found that directly fine-tuning T5-Base with these modifications on downstream tasks led to very poor results and do not include them here. Instead, we train T5-Set (analogous to Set) for a fair point of comparison.

We report a subset of the results from the main paper in for T5-Base in \cref{tab:within-pretrain-t5,tab:bimachines-t5,tab:g2p-t5}.

We also tried to pre-train a ByT5-style model from scratch (i.e. from random initialization). However, we could not find a setting of hyperparameters that would make the model converge well. We hypothesize that the model already needs to be in a reasonable space to make learning feasible.

\begin{table*}[t]

\begin{minipage}[t]{0.51\linewidth}
\centering

\begin{tabular}{lrrrr}
\toprule
& \multicolumn{2}{c}{Iteration} & \multicolumn{2}{c}{UC} \\
  & Acc$\uparrow$ & ED$\downarrow$ & Acc$\uparrow$ & ED$\downarrow$ \\
\midrule
T5-Set & 26.6 & 6.26 & 55.1/54.6 & 1.18/1.02 \\
T5-SIP-d4 & 94.5 & 0.11 & 75.4/99.5 & 0.54/0.01 \\
\bottomrule
\end{tabular}
\caption{Evaluating systematic generalization on FST tasks with 4 states (cf. \cref{tab:within-pretrain}). Due to an outlier task on UC, we additionally report the median after `/'.}
    \label{tab:within-pretrain-t5}
\end{minipage}
\hfill
\begin{minipage}[t]{0.44\linewidth}
\centering

\begin{tabular}{lrrrr}
\toprule
 & \multicolumn{2}{c}{Iteration} & \multicolumn{2}{c}{UC} \\
 & Acc$\uparrow$ & ED$\downarrow$ & Acc$\uparrow$ & ED$\downarrow$ \\
\midrule
T5-Set & 77.9 & 0.73 & 81.7 & 0.53 \\
T5-SIP-d4 & 83.3 & 0.56 & 86.1 & 0.37 \\
\bottomrule
\end{tabular}
    \caption{Evaluation with T5-Base on non-deterministic FSTs (cf. \cref{tab:bimachines})}
    \label{tab:bimachines-t5}
\end{minipage}

\end{table*}

\begin{table*}[t]
    \centering

\addtolength{\tabcolsep}{-0.38em}

\begin{tabular}{lrrrrrrrrrrrrrr|rr}

\toprule
 & \multicolumn{2}{c}{ban} & \multicolumn{2}{c}{cop} & \multicolumn{2}{c}{got} & \multicolumn{2}{c}{lao} & \multicolumn{2}{c}{syl} & \multicolumn{2}{c}{tel} & \multicolumn{2}{c}{tzm} & \multicolumn{2}{c}{Avg} \\
 & Acc & PER & Acc & PER & Acc & PER & Acc & PER & Acc & PER & Acc & PER & Acc & PER & Acc$\uparrow$ & PER$\downarrow$ \\
\midrule
T5-Set & 47.9 & .231 & 1.2 & .783 & 6.7 & .458 & 3.6 & .643 & 6.6 & .611 & 4.9 & .612 & 2.7 & .797 & 10.5 & .591 \\
T5-SIP-d4 & 59.1 & .154 & 4.7 & .640 & 69.6 & .059 & 5.9 & .566 & 22.1 & .447 & 35.4 & .191 & 12.5 & .509 & 29.9 & .367 \\
\bottomrule
\end{tabular}
\caption{Grapheme-to-phoneme conversion with 100 training examples based on T5-Base. In contrast to the experiments in the main paper, we found that T5-SIP-d4 did not perform well on completely unseen scripts, so we mapped all Unicode code points to arbitrary ASCII characters. This maintains the structure of the task and is completely reversible. T5-Set is evaluated in the same way.}
    \label{tab:g2p-t5}
    
\addtolength{\tabcolsep}{0.38em}

\end{table*}

\subsection{Generalization to longer strings}
\begin{table*}[t]
    \centering

    \begin{tabular}{lllrr}
    \toprule
    Test length & Model & pre-train length &  Acc$\uparrow$ & ED$\downarrow$ \\
    \midrule
    \multirow[t]{2}{*}{40 to 70} & ByT5 & 1024 & 29.3 & 15.60 \\
     & \dfourmethod & 35 & \textbf{69.4} & \textbf{2.61} \\ [1.0ex]
    \multirow[t]{3}{*}{90 to 110} & ByT5 & 1024 & 1.4 & 55.37 \\
     & \dfourmethod & 35 & 3.4 & 34.50 \\
     & \dfourmethod-long & 110 & \textbf{81.5} & \textbf{1.09} \\
    \bottomrule
    \end{tabular}
    \caption{Average generalization ability across 5 FSTs with 4 states. Models were trained on inputs of length up to 15, and tested on much longer inputs.}
    \label{tab:longer-strings}
\end{table*}

In the main paper, we report results on iteration generalization where a model is trained on strings such that each state has been visited at most 3 times, and is tested on strings where at least one state is visited at least 4 times. Here, we explore a more extreme version, where there is a large gap between the maximum length seen during training and the minimum length seen during testing. As another point of comparison, we further pre-train \dfourmethod on 40,000 FSTs with strings of length up to 110 (\dfourmethod-long).

We report results in \cref{tab:longer-strings}. ByT5 struggles with this generalization setup across the board. \dfourmethod performs remarkably well on lengths 40-70 which are beyond the lengths seen during its pre-training. However, performance drops starkly when testing on inputs of length 90 to 110. We hypothesize that this is because the relevant positional embeddings were not pre-trained by \method. In contrast, \dfourmethod-long performs well on inputs of length 90 to 110, as it has seen strings of such length during pre-training.

\section{Hallucination Example}
\label{appendix:hallucination}
We briefly show an example where an LLM ignores a part of the input and resorts to outputting a high-frequency entity. Consider the following in-context examples for a simple text editing task:
\begin{center}
\begin{tabular}{ll}
\textbf{Input}	& \textbf{Output} \\
Howard Phillips Lovecraft	& H.P. Lovecraft \\
John Ronald Reuel Tolkien	& J.R.R. Tolkien \\
Thomas Stearns Eliot	& T.S. Eliot \\
\end{tabular}
\end{center}
At the time of submission, the current version of ChatGPT frequently outputs ``J.K. Rowling'' for the name ``John Edward Rowling'', hallucinating the K.

\section{Additional Analysis}

\subsection{Analysis of fine-tuned prefixes}
\label{appendix:analysis-prefix}
To gain some understanding of how the prefix of tunable embeddings is used by the model and what it contains, we consider the setup of fine-tuning only the prefix and keeping the rest of the model unchanged. That is, all the task-specific information has to be captured in these embeddings. Specifically, we fine-tune on the 5 FSTs from \cref{sec:within-pretrain} for iteration generalization for 20 epochs with a learning rate of 0.5. 

We explore two questions: 
\begin{enumerate}
    \item Is the model robust towards different permutations of the fine-tuned prefixes? Intuitively, these permutations correspond to changing the order in which transitions are listed, so ideally the model should not be sensitive to that order.
    \item Does the fine-tuned prefix represent the task-specific information in a similar way to how FSTs were encoded during pre-training?
\end{enumerate}
To address the first question, we randomly permute the tuned prefixes and compute accuracy on the iteration generalization data before and after permuting the tuned prefixes. We use 20 permutations per learned prefix and average results across the 5 FSTs. Overall, we find that this results only in a small drop in accuracy: the median drop in accuracy is only around 1.3 percentage points, and the arithmetic mean of the drop is around 7.1 percentage points. Most permutations do not have a big impact on how the prefix is interpreted but a few permutations do have a stronger negative impact, skewing the arithmetic mean.

\begin{figure}[t]
    \centering
    \includegraphics[width=\linewidth]{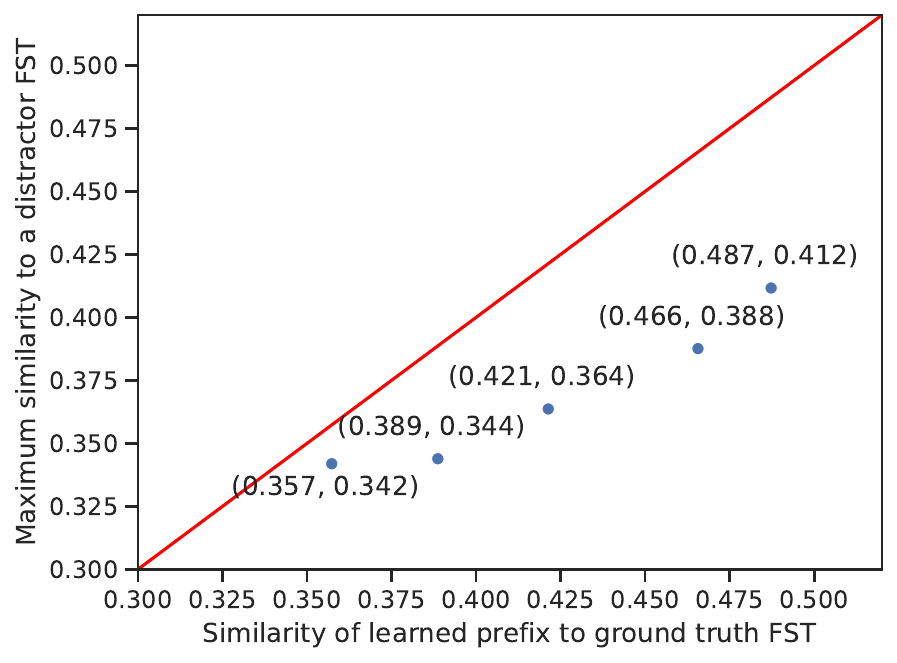}
    \caption{Each dot represents a fine-tuned prefix when the rest of the model remains frozen during fine-tuning. The x-coordinates represent the similarity to a ground truth gold prefix, and the y-coordinates represent the maximum similarity to any of the $5 \times 10000$ distractor FSTs. All dots are below the diagonal, hence all learned prefixes are most similar to an encoding of the ground truth FST.}
    \label{fig:similarities-learned-prefixe}
\end{figure}

To address the second question, we test if the learned prefix for a task $t$ resembles an encoding of an FST that solves $t$. For each of the 5 FSTs, we generate 10,000 distractors, \ie FSTs that have the same number of states and use the same vocabulary as the FST solving $t$.
We define the similarity of two prefixes $p, q$ as follows:
$$sim(p, q) = \max_{\pi} \frac{1}{n} \sum_i \frac{p_i^T q_{\pi(i)} } {||p_i||_2 \cdot ||q_{\pi(i)}||_2 }$$
where $\pi$ is a permutation, and $p_i$ is the $i$-th vector in prefix $p$, and prefixes $p$ and $q$ both have length $n$. That is, we define the similarity between $p$ and $q$ as the highest possible average cosine similarities between positions in $p$ and $q$ that one can achieve by assigning a position in $p$ to exactly one position in $q$ and vice-versa.\footnote{Computing the similarity $sim(p, q)$ is relatively expensive because it involves solving the assignment problem (\eg with the Hungarian algorithm). Instead of solving the assignment problem exactly, we approximate it with the Sinkhorn algorithm \citep{sinkhorn}. We then take the output of the algorithm (a matrix of `soft' assignments) and for each position in $p$, we greedily select a matching position in $q$.} Taking the maximum over all permutations is justified by our results to the first question above, which showed that the model is largely invariant to different permutations of the tuned prefix.

For every task $t$, we compute the similarity between the prefix $p$ learned by fine-tuning on input/output pairs and the union of encodings of the distractors and encodings of the gold standard FST for task $t$. Where necessary, we truncate encodings of FSTs to have the same length as the learned prefix. We present the results in \cref{fig:similarities-learned-prefixe} showing that all learned prefixes are most similar to an encoding of the ground truth FST.

\subsection{Probing non-\method models}
\label{appendix:probe-non-sip}

All probes are trained for one epoch on activations produced by passing 8,000 FSTs with 5 inputs each (\ie 40,000 instances) through the model.

For the baseline probe, we take the trained \dfourmethod model (including matrix $W$ and embeddings from \ref{sec:pre-train}) and re-initialize the Transformer to ByT5-small. The probe achieves only a token-level accuracy of 42.9\% and whole-sequence accuracy of 8.1\%. We see very similar results for a probe trained on a randomly initialized Transformer in this setup: a token-level accuracy of 42.5\% and a whole-sequence accuracy of 7.1\%.

\section{Additional model details \& Hyperparameters \& Hardware}
\label{appendix:additional-model-details}

\paragraph{\method}
For completeness, we now describe the order in which we arrange the transitions. While the ordering of the transitions does not matter for expressing FSTs, the Transformer uses positional encodings which might have impacts on the pre-training (though see \cref{appendix:analysis-prefix}).
We assemble the overall prefix by stacking the individual vectors $h_0, \ldots ,h_n$ of the transitions $p_0 \trans{\sigma_0}{\rho_0} q_0, \ldots, p_n \trans{\sigma_n}{\rho_n} q_n $. We group the transitions by their originating state (\ie $p_i$) and go over the states by their id, starting with 0, the initial state.

During pre-training, we might encounter FSTs with different numbers of transitions within the same batch. To handle this, we use padding encodings by reserving a special padding state and padding symbol in the embedding matrices of states and symbols. To initialize the prefix for fine-tuning, we use the average of 32 FST encodings (chosen at random) from pretraining.

For pre-training, we use embeddings of dimensionality 64 for states, embeddings of dimensionality 256 for symbols, and of dimensionality 16 to indicate final/non-final states.

\paragraph{Task embeddings} To enable faster adaption of the task embeddings than the rest of the model to fit a particular task, we use a higher learning rate for the task embeddings (1.0) than for the rest of the model ($5 \cdot 10^{-4}$) during pre-training. We also use a higher learning rate for the prefix during fine-tuning, analogously to \method. 

Because we have to store 40,000 task embeddings (one for each generated FST), TE requires a lot of memory. To reduce memory consumption, the task embeddings have a dimensionality of 180 and are up-projected to fit into the Transformer, analogously to W in \cref{sec:pre-train}.
Nevertheless, the memory consumption of the embeddings is substantial and we store them on a separate GPU. Analogously to \dfourmethod, we pre-train for 20 epochs.

\paragraph{Naive} We pre-train for a single epoch only as we found this achieved better results on downstream tasks than training for 20 epochs.

\paragraph{Set} We sample 200,000 examples according to the procedure described by \citet{wu2022insights} to match our pre-training dataset size. Again, we found it more helpful for downstream task performance to train for a single epoch rather than 20 epochs.

\paragraph{Fine-tuning Hyperparameters} Like pre-training, we finetune with the Adam optimizer. The main hyperparameters involved for both \method and TE are the learning rates for the main model, and (separately) the learning rate of the tunable prefix. We chose these manually. Generally, we found that using a learning rate of $1.0$ was a good choice for the prefix. \citet{lester-etal-2021-power} report a similarly high learning rate to be useful for prompt tuning. 
For the rest of the model, we found $3 \cdot 10^{-4}$ and $5 \cdot 10^{-4}$ to work well for \dfourmethod and TE, respectively. For few-shot experiments, we use a somewhat smaller learning rate for TE for the main model ($3 \cdot 10^{-4}$). We noticed that T5-SIP-d4 (see \cref{appendix:t5-base}) was more sensitive to the learning rate choice in general than \dfourmethod.

\paragraph{Hardware/Computational Budget} We ran our experiments on NVIDIA GeForce RTX 2080 Ti GPUs (11264MiB RAM) with driver version 535.54.03 and cuda version 12.2. 

Pre-training \dfourmethod took around 30 hours for 20 epochs. One training run on synthetic data (including evaluation) takes around one hour, and one training run for low-resource grapheme-to-phoneme conversion takes between 5 and 10 minutes.

\end{document}